\documentclass[conference]{IEEEtran}

\usepackage{ifpdf}
\usepackage{cite}
\usepackage[pdftex]{graphicx}
\usepackage[cmex10]{amsmath}
\usepackage{algorithmic}
\usepackage{array}
\usepackage{mdwmath}
\usepackage{mdwtab}
\usepackage{eqparbox}
\usepackage[tight,footnotesize]{subfigure}
\usepackage{stfloats}
\usepackage{url}
\usepackage{amssymb}
\usepackage[hidelinks]{hyperref}

\begin{document}
%
\title{Automatic Script Identification in the Wild}
%
%
%

\author{
\IEEEauthorblockN{Baoguang~Shi$^{*}$, Cong~Yao$^{*}$, Chengquan~Zhang$^{*}$, Xiaowei~Guo$^{\dag}$, Feiyue~Huang$^{\dag}$, Xiang~Bai$^{*}$}
\IEEEauthorblockA{$^{*}$School of EIC, Huazhong University of Science and Technology, Wuhan, P.R.~China 430074 \\
$^{\dag}$Tecent, Shanghai, P.R.~China 200233\\
$^{*}${\small \{shibaoguang, yaocong2010, zchengquan\}@gmail.com} $^{\dag}${\small \{94631283, 34115813\}@qq.com} $^{*}${\small xbai@hust.edu.cn}
}}

\maketitle

\IEEEpeerreviewmaketitle

\begin{abstract}
With the rapid increase of transnational communication and cooperation, people frequently encounter multilingual scenarios in various situations. In this paper, we are concerned with a relatively new problem: script identification at word or line levels in natural scenes. A large-scale dataset with a great quantity of natural images and 10 types of widely-used languages is constructed and released. In allusion to the challenges in script identification in real-world scenarios, a deep learning based algorithm is proposed. The experiments on the proposed dataset demonstrate that our algorithm achieves superior performance, compared with conventional image classification methods, such as the original CNN architecture and LLC.\footnote{This paper has been submitted to International Conference on Document Analysis and Recognition (ICDAR) 2015 and is now under peer-review.}
\end{abstract}

%
\IEEEpeerreviewmaketitle

\section{Introduction} \label{sec:introduction}

Automatic script identification is a task that facilitates many important applications in both document analysis and natural scene text recognition. Previous work mostly focuses on script identification in documents~\cite{DBLP:journals/pami/Tan98, DBLP:journals/pami/HochbergKTK97, DBLP:journals/pami/GhoshDS10} or videos~\cite{DBLP:conf/icdar/PhanSDLT11, DBLP:conf/das/ZhaoSLT12}. In document, script identification can be done at page/paragraph level~\cite{DBLP:journals/ijdar/JoshiGS07}, text line level~\cite{DBLP:conf/icdar/PalSC03}, word level~\cite{DBLP:conf/das/SinhaPC04} or character level~\cite{DBLP:conf/icdar/RaniDL13}. Tan~\cite{DBLP:journals/pami/Tan98} investigated the properties of a group of rotation invariant texture features and used these features to recognize the language type of characters in machine printed document. Hochberg~\emph{et al.}~\cite{DBLP:journals/pami/HochbergKTK97} propose a script identification system for characters stored electronically in image form.

\begin{figure}[t]
    \begin{centering}
   \includegraphics[width=0.9\linewidth]{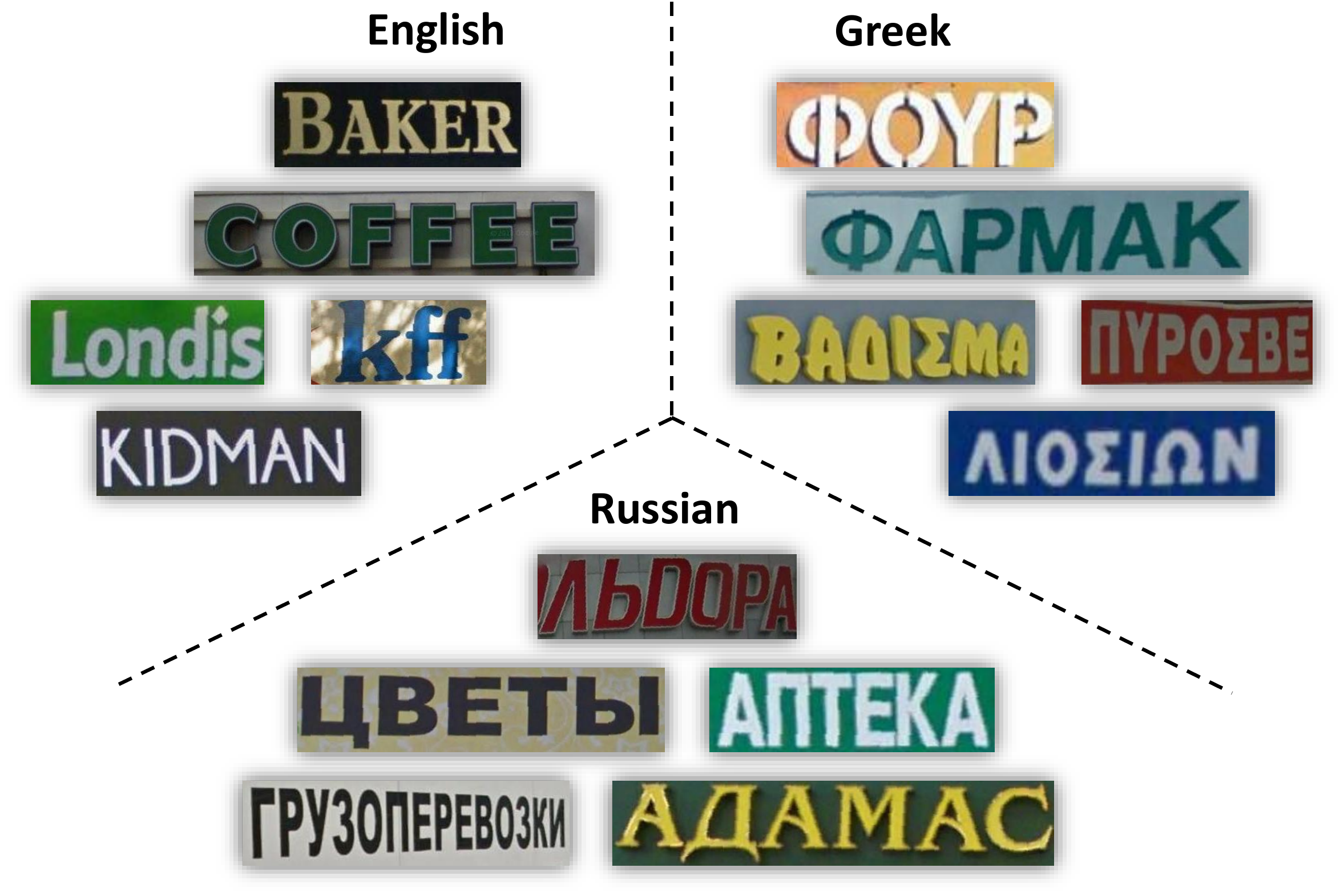}
    \par \end{centering}
    \caption{Illustration of scripts that share subsets of alphabets. Characters ``A'', ``B'' and ``E'' appear in all these three scripts. The script identification relies on special characters that are unique to particular scripts.}
    \label{fig:sharedAlphabet}
\end{figure}

In this paper, we are concerned with a relatively new problem: script identification at word or line levels in natural scenes. Identifying script in natural scene is an important task, particularly to text reading systems under multilingual scenarios~\cite{DBLP:conf/icdar/GomezK13}. Naturally, this problem can be casted as an image classification problem, which has been studied extensively~\cite{DBLP:conf/cvpr/WangYYLHG10, DBLP:conf/nips/KrizhevskySH12}. Nevertheless, it remains a challenging problem, mainly due to four reasons: (1) Characters in the wild are usually with higher variability in font, color and layout. (2) Backgrounds in natural scenes are more complex and may contain clutter or noise. (3) Different scripts may share a subset of alphabets, as illustrated in Figure~\ref{fig:sharedAlphabet}. This phenomenon makes it difficult to distinguish among different types of languages solely from appearance. (4) Text images are in arbitrary lengths, ruling out some classification methods that only operate on fixed sized inputs. To deal with the challenges, we propose a deep leaning based unified framework to recognizing scripts in the wild. Towards this end, we make the following contributions:
\begin{itemize}
\item We establish a large-scale benchmark for algorithm development and comparison. The benchmark includes 13045 word images, cropped from 7700 full images taken in diverse real-world scenarios. The scripts in the images are from 10 different languages.
\item To tackle the challenges described above, we propose a deep leaning based algorithm, which could serve as the baseline algorithm in future research.
\item Compared with other conventional image classification methods, our approach better exploits the characteristics of texts in natural images and obtains superior performance when evaluated on the proposed benchmark.
\end{itemize}

\section{The SIW-10 Dataset} \label{sec:dataset} 
\begin{figure}[t]
    \begin{centering}
   \includegraphics[width=0.9\linewidth]{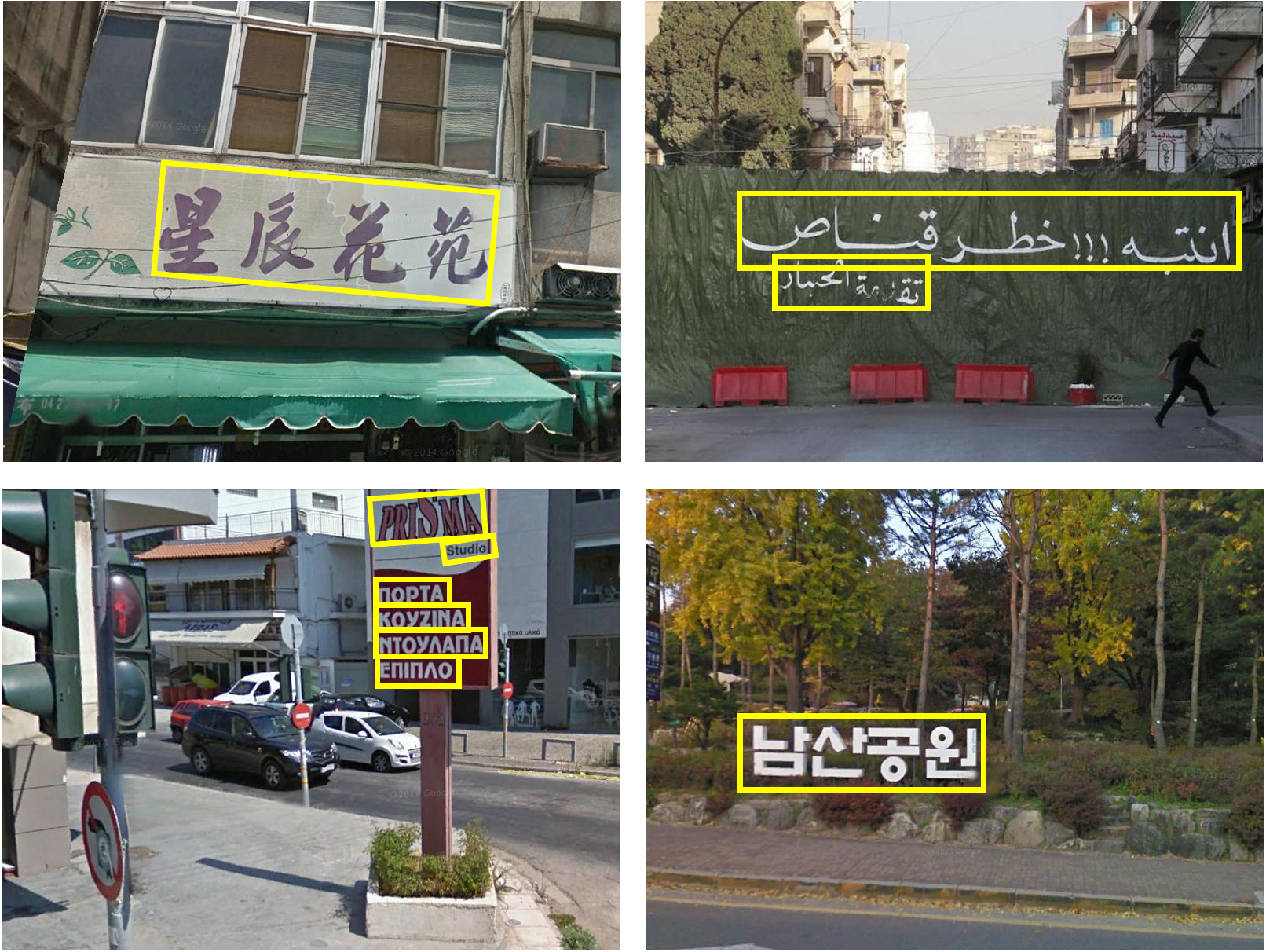}
    \par \end{centering}
    \caption{Examples of images that we harvested from Google Street View, along with the annotations.}
    \label{fig:fullImage}
\end{figure}

\begin{figure*}[t]
    \begin{centering}
   \includegraphics[width=1.0\linewidth]{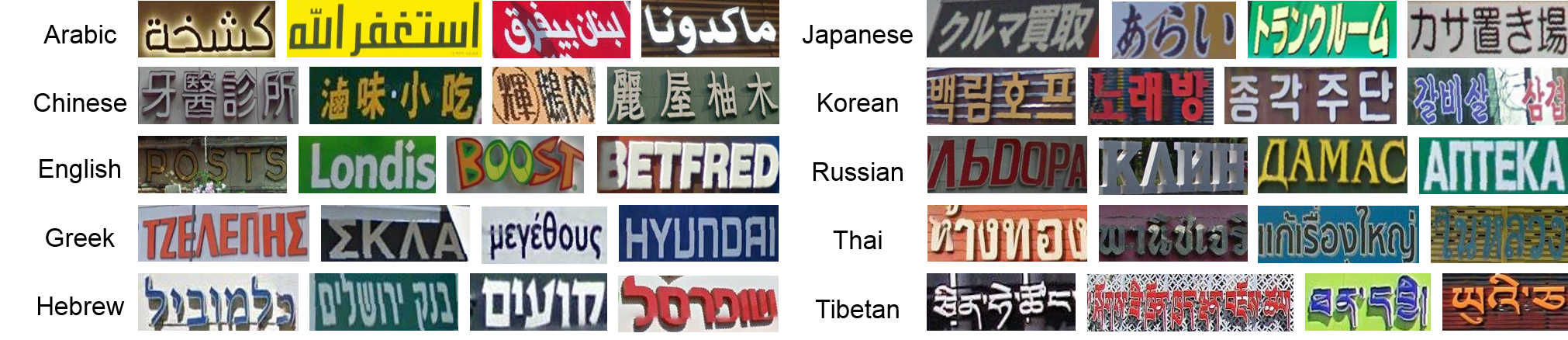}
    \par \end{centering}
    \caption{Some examples in the SIW-10 dataset. The dataset contains all together 13,045 cropped images of words or text lines in 10 classes.}
    \label{fig:dataExample}
\end{figure*}

There exist several public datasets that consist of natural images with texts, for instance, ICDAR 2011~\cite{DBLP:conf/icdar/ShahabSD11a}, SVT~\cite{DBLP:conf/eccv/WangB10} and IIIT 5K-Word~\cite{DBLP:conf/bmvc/MishraAJ12}. However, these datasets are primarily used for scene text detection and recognition tasks. Besides, they are dominated by English or other Latin-based languages. In the area of script identification, there exists datasets~\cite{DBLP:journals/pami/HochbergKTK97, DBLP:conf/icdar/PhanSDLT11, DBLP:conf/das/ZhaoSLT12}. But these databases only include characters in document images or videos, rather than natural images.

Therefore, we propose a new dataset for script identification in wild scenes\footnote{We will release the dataset (cropped images and full sized images) for academic use.}. The dataset contains multi-scripts  images that are taken from natural scenes images (Figure~\ref{fig:fullImage}). As illustrated in Figure~\ref{fig:dataExample}, the database includes text images from 10 languages: Arabic, Chinese, English, Greek, Hebrew, Japanese, Korean, Russian, Thai and Tibetan. Hence, we call this benchmark Script Identification in the Wild 10 Classes (SIW-10) dataset.

We first harvest a collection of street view images from the Google Street View and manually label the text regions by their bounding boxes (Figure ~\ref{fig:fullImage}). Text line images are then cropped from these images. For each language category, 600 to 1,000 street view images are collected and 1,000 to 2,000 text regions are extracted. In our dataset, we include only horizontal text images that contain one or several words. The dataset contains 13,045 cropped images of words or text lines. Among them 5,000 are used for testing and the rest 8,045 are used for training.

Note that the SIW-10 dataset is diverse and challenging. The images are captured from different locations all over the world, under different imaging conditions. Majority of the images are with low resolution, noise or blur. We believe the SIW-10 dataset can be serve as a standard benchmark for script identification in the wild.

\section{Our Approach} \label{sec:approach}

Considering the challenges discussed in Section~\ref{sec:introduction}, it would be desirable if our approach is able to capture distinctive characters or components in the text images, and able to deal with inputs with arbitrary aspect ratios.

Recent research on image classification has seen a leap forward, thanks to the wide applications of deep convolutional neural networks (CNNs, \cite{lecun1998gradient}). CNNs are designed to operate on input maps with fixed widths and heights. Text images, however, come in arbitrary sizes and aspect ratios. Their aspect ratios vary greatly, depending on the number of characters they contain.

To utilize the power of automatic feature learning in CNNs, while making it fit to the script identification problem, in the following we propose Multi-stage Spatially-sensitive Pooling Network (MSPN), a novel variant of CNN, for the script identification task. The network efficiently captures rich and distinctive features in text images for script identification, while inherently and naturally deals with input images with arbitrary aspect ratios.

\begin{figure*}[t]
    \begin{centering}
    \includegraphics[width=0.9\linewidth]{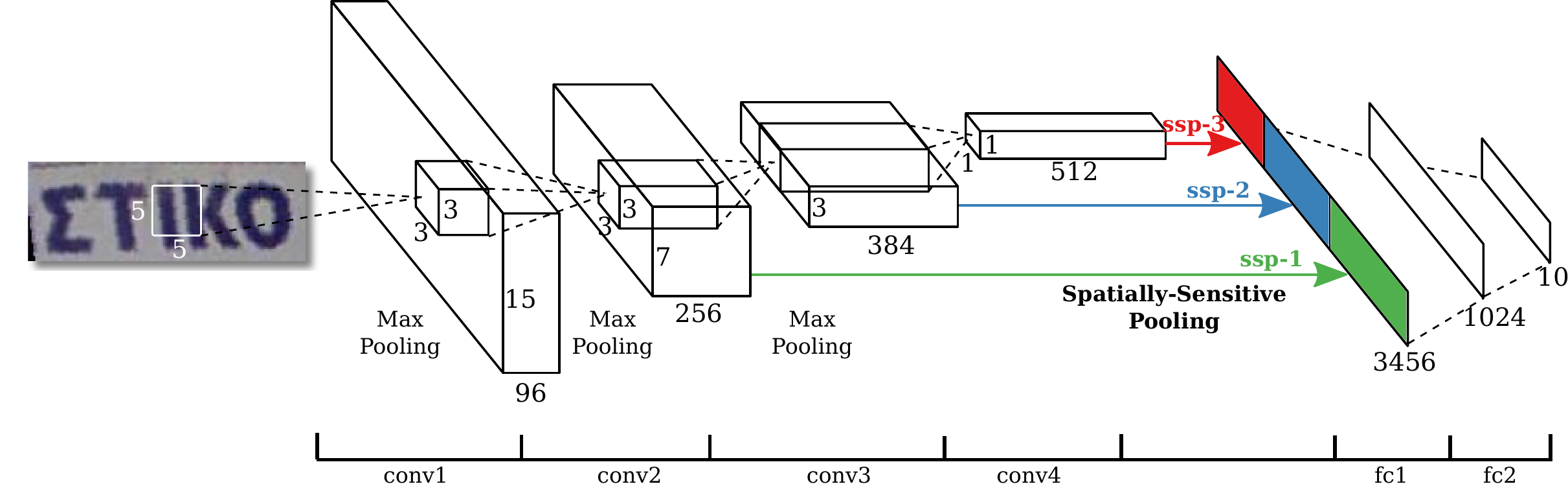}
    \par \end{centering}
    \caption{MSPN architecture illustrated following the style used in \cite{DBLP:conf/nips/KrizhevskySH12}. The four cuboids after convolutional stages represent response maps. The size for each of the cuboids corresponds to map width $\times$ map height $\times$ number of maps. The cuboids inside represent convolutional kernels. The spatially-sensitive pooling layers are indicated by arrows in three different colors. The output is the softmax probability vector of length 10. See Section~\ref{sec:arch} for details.}
    \label{fig:architecture}
\end{figure*}

\subsection{Architecture} \label{sec:arch}

The architecture of the network is depicted in Figure~\ref{fig:architecture}. As the preprocessing step, the input images are resized to have fixed height ($32$ pixels throughout our experiments), keeping their aspect ratios. The first four convolutional layers (with max-pooling and rectifier layers) in the network work the same way as they do in a CNN, except that the sizes of the input maps are arbitrary. Since convolutional layers apply filter banks to all places in an image, these layers are inherently capable of dealing with images in arbitrary sizes. The sizes of their output maps change with the input sizes. In our network, images are fixed in height. Therefore the response maps produced by these convolutional layers are fixed in height but varied in length. In our settings, output maps have heights $15$, $7$, $3$, $1$ respectively and widths proportional to the width of the input image. These layers aim to capture rich, hierarchical features from raw image pixels.

In a conventional CNN structure, response maps output by the last convolutional layer are flattened and fed to the following fully-connected layers or locally-connected layers. These kinds of layers can only accept inputs with fixed number of dimensions. They are not suitable for our problem due to that text images vary in lengths. Besides, the discriminative features in a text image can appear at any horizontal positions, since that characters are arranged in different orders, but their vertical positions still matter. To address these problems, we propose the spatially-sensitive pooling layer, which accepts input maps with arbitrary widths, and captures topological information along vertical directions.

\subsection{Spatially-sensitive pooling layer}

As have discussed in Section~\ref{sec:dataset}, distinctive parts play an important role in script identification. Typically, text is a collection of characters arranged in a line. A distinctive character or character component may appear at any horizontal position in the image, so that their horizontal positions are less informative. But their vertical positions still matter as characters are written upright. On the other hand, text images are in arbitrary aspect ratios. Conventional CNNs cannot deal with them directly. We propose the Spatially-Sensitive Pooling (SSP) layer, which captures useful features for script identification, as well as deals with arbitrary input sizes.

The SSP layer takes input maps that have a fixed number of rows but a variable number of columns. For each of the input maps, the SSP layer pools along each row of the map by taking the maximum or average value in each row. Assuming that the input is a tensor of sizes $n_{\mathrm{map}} \times w \times h$, where $n_{\mathrm{map}}$ is the number of input maps, $w$, $h$ are the width and height of the maps. Then the output would be a vector of length $n_{\mathrm{map}} h$, which is independent on the width of the input images.

The SSP layer introduces invariance to horizontal positions of responses to the network. Meanwhile it keeps vertical positions of the responses. This makes it suitable for describing images of texts in a line. Furthermore, SSP layers accept input maps with arbitrary widths and output vectors with fixed lengths, thus capable of acting as the bridge between the convolutional layers and fully-connected layers. As a consequence, the network is able to deal with images with arbitrary sizes and aspect ratios inherently and naturally.

\subsection{Multi-stage pooling}

The distinctive features could be at different abstraction levels. Both higher level features and lower level features may help the recognition process. To utilize features at different abstraction levels, we introduce a multi-stage pooling scheme into the network. As illustrated in Figure~\ref{fig:architecture}. The colored lines whose arrowheads start from \textbf{conv2}, \textbf{conv3} and \textbf{conv4} layers indicate SSP layers that are inserted after these convolutional layers. The output of the three pooling layers are concatenated as a long vector, which is fed to later fully-connected layers. Thereafter, the pooling features in the long vector contain pooling features from all the three SSP layers. Since layers \textbf{conv2}, \textbf{conv3} and \textbf{conv4} output response maps at different abstraction levels, the features concatenated from their pooling features describe the text image in both high-levels and low-levels. The resulting features are rich and describe multiple aspects of the text image, which is desirable for script identification.

The multi-stage pooling scheme results in a graph structured network that is more complex than simple sequential structured network. In this network, convolutional layers take errors back-propagated from multiple layers. For example, \textbf{conv2} receives errors back-propagated from both \textbf{conv3} and \textbf{fc1}. These convolutional layers (\textbf{conv2} and \textbf{conv3} in our network) are trained with respect to gradients on pooling features as well as response maps. This potentially encourages these layers to produce more discriminative response maps, due to that their outputs are directly used for classifying after pooling. Therefore, the network benefits from the the rich and discriminative features.

\section{Experiments} \label{sec:experiments}

We evaluate our MSPN on the collected SIW-10 dataset. Besides, we implement a baseline algorithm using the conventional CNN, with some simple workarounds to bypass the variable aspect ratio problems. As another baseline, we implement the Locality-constrained Linear Coding (LLC)~\cite{DBLP:conf/cvpr/WangYYLHG10} algorithm, which is widely used in image classification tasks. We compare the performances of these approaches with the proposed approach on the SIW-10 dataset.

\subsection{Dataset and Implementation Details}

\subsubsection{Dataset}
We evaluate our algorithms on the SIW-10 dataset. We build the testing set with all together $5,000$ images with $500$ testing images for each class. The rest $8,045$ images are for training. The number of training images for all the script classes are respectively: Arabic 503, Chinese 809, English 725, Greek 522, Hebrew 770, Japanese 717, Korean 1064, Russian 532, Thai 1726 and Tibetan 677. For all classes the number of testing samples are 500.

\subsubsection{Implementation details} \label{sec:implDetails}
We train the network using stochastic gradient descent (SGD)~\cite{DBLP:series/lncs/LeCunBOM12} with the initial learning rate set to $0.01$ and the momentum set to $0.9$. Following~\cite{DBLP:conf/nips/KrizhevskySH12}, the learning rate is decreased by a factor of $0.1$ when the validation error plateaus. The training terminates once the learning rate is less than $1\times 10^{-5}$.

\subsection{Baseline Methods}

\subsubsection{CNN-Simple}

We setup a convolutional neural network to identify the scripts from text images. The CNN structure we use is similar to our MSPN in the convolutional parts. CNN cannot deal with inputs with arbitrary sizes. To address this problem, we first sample patches in the input images and resize them to a fixed size, as illustrated in Figure~\ref{fig:architecture}. Patches are set to labels that are same to the image they are cropped from. In our implementation, text images are first resized to have heights of $40$ and squared patches with sizes randomly chosen within range $[25, 40]$. All the patches are then resized to $32 \times 32$ after sampling. The CNN is trained on the patches sampled from training images, with the same scheme described in Section~\ref{sec:implDetails}.

During the testing process, we first predict the labels of patches using the trained CNN. Their classification probability vectors $\mathbf{p}$ output by the CNN soft-max layer are then used to predict the class of the whole text image. Denote the set of patches cropped from text image $I^{(i)}$ by $\{ \mathbf{x}_{j}^{(i)} \}_{j=1}^{n_{i}}$, where $n_{i}$ is the number of patches sampled from image $I^{(i)}$. The prediction is done by:
\begin{equation}
    \mathbf{p}^{(i)}_{j} = \mathrm{CNN} (\mathbf{x}_{j}).
\end{equation}

In order to get the final prediction on the whole image, we combine the predictions on all patches by calculating the average of their probabilities:
\begin{equation}
    \mathbf{y}^{(i)} = \frac{1}{n_{i}} \sum_{j=1}^{n_{i}} \mathbf{p}^{(i)}_{j}.
\end{equation}

The resulting $\mathbf{y}^{(i)}$ is taken as the multi-class score vector for text image $I^{(i)}$. The prediction is made by choosing the class with the max score in $\mathbf{y}^{(i)}$.

\subsubsection{LLC}

The LLC~\cite{DBLP:conf/cvpr/WangYYLHG10} is a widely adopted image classification algorithm. It is based on the Bag-of-Words model (BoW), max-pooling and spatial pyramid matching (SPM)~\cite{DBLP:conf/cvpr/YangYGH09}. We densely sample SIFT descriptors at $3$ different scales. A part of them are used to build a codebook with $2048$ codewords. Training and testing images are coded using the LLC coding scheme. The SPM is applied by vertically dividing text images into respectively $2$ and $3$ subregions with equal heights. The resulting coding features are in $(1+2+3)\times 2048 = 12288$ dimensions. Finally, a linear SVM~\cite{DBLP:journals/ml/CortesV95} classifier is learned on the coding features.

\subsection{Evaluation}

\begin{figure}[t]
    \begin{centering}
   \includegraphics[width=\linewidth]{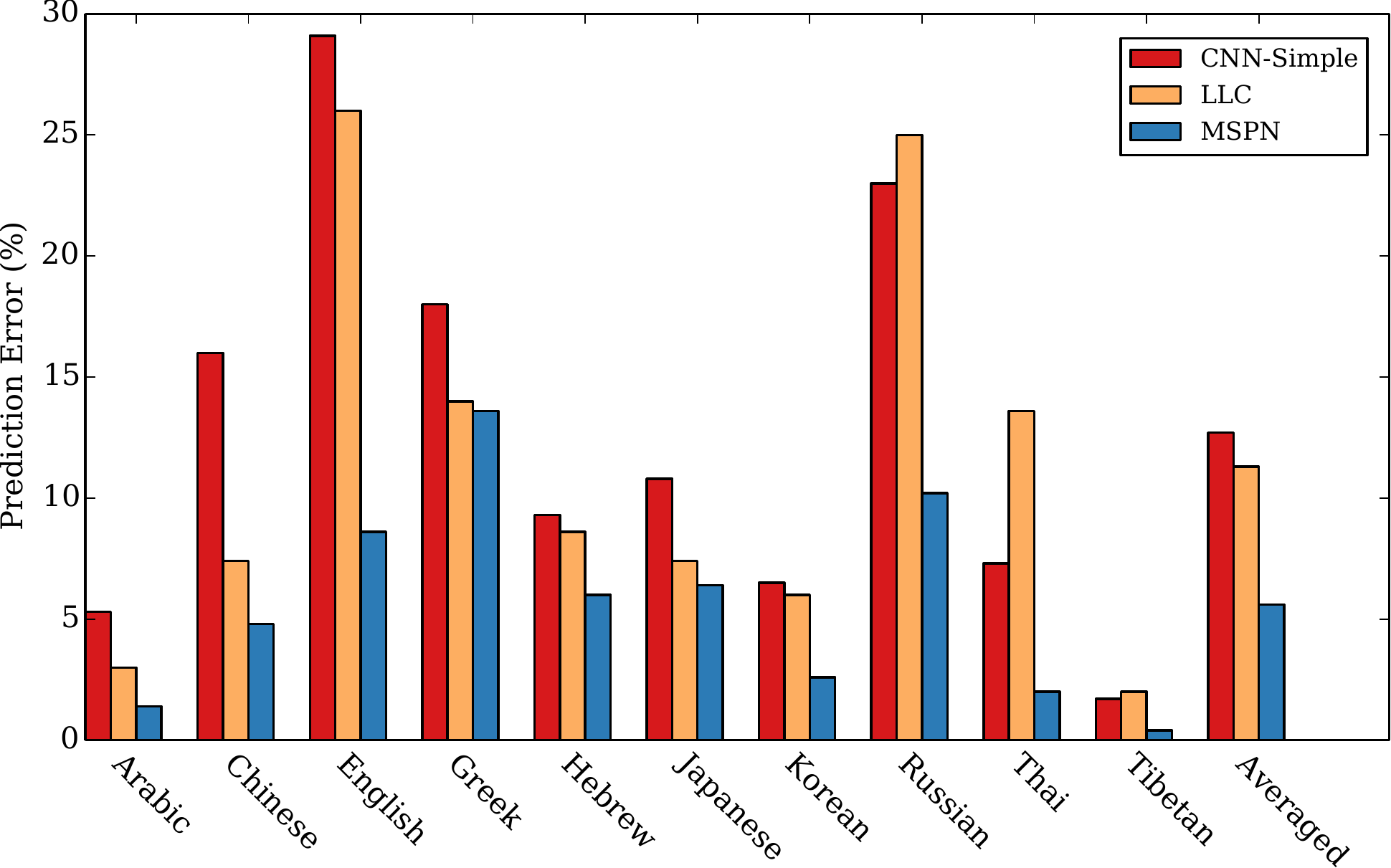}
    \par \end{centering}
    \caption{Comparisons on recognition accuracies among CNN-Simple, LLC and MSPN, evaluated on the SIW-10 dataset.}
    \label{fig:siwCompare}
\end{figure}

We evaluate our MSPN as well as the baseline methods on the SIW-10 dataset. The prediction accuracies for each class is evaluated and compared in Figure~\ref{fig:siwCompare}. It can be seen that our approach achieves the best results on all classes and reduces the error of LLC by a large margin.

\begin{figure}[t]
    \begin{centering}
    \includegraphics[width=0.8\linewidth]{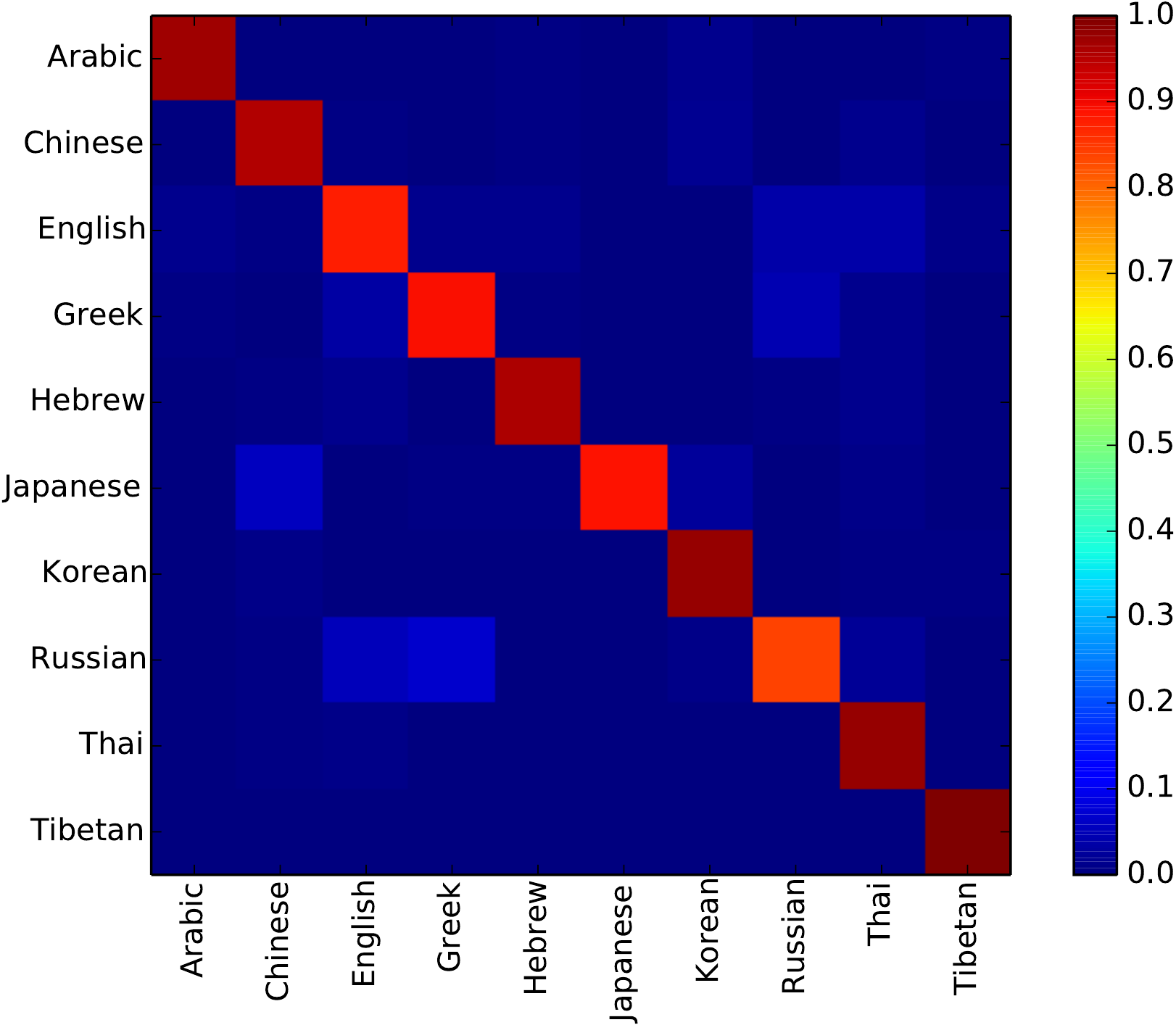}
    \par \end{centering}
    \caption{Prediction confusion matrix on SIW-10 made by MSPN. On Y-axis are ground truth labels and on X-axis are predictions. Mis-classifications frequently happen between Chinese and Japanese; Russian and Greek; Russian and English, \emph{etc.}}
    \label{fig:confMat}
\end{figure}

\begin{figure}[t]
    \begin{centering}
    \includegraphics[width=0.7\linewidth]{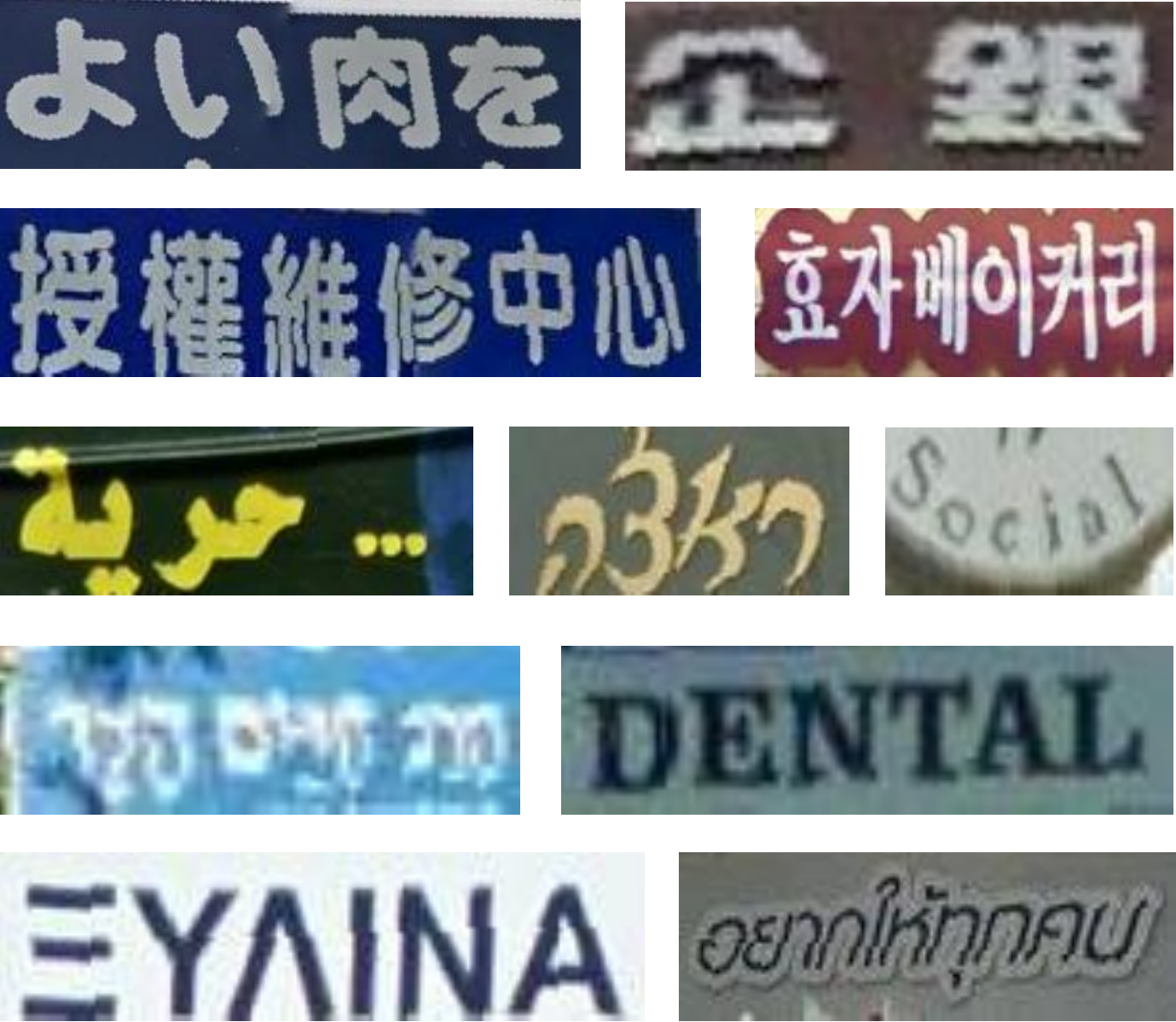}
    \par \end{centering}
    \caption{Some misclassified samples. The misclassification are mainly due to the shared characters between languages (e.g. Japanese vs. Chinese), unusual layout and cluttered background.}
    \label{fig:failureCases}
\end{figure}

From the confusion matrix shown in Figure~\ref{fig:confMat} we can see that mis-classifications frequently happen between several pairs of scripts, e.g. Chinese and Japanese. The reason is most likely to be that these languages share a large proportion of alphabets. Some words are even indistinguishable without semantic information, which we have not yet incorporated into our framework.

The CNN-Simple approach performs significantly worse than the MSPN. The reasons could be that 1) in CNN-Simple, the network is not directly optimized with respect to a loss function that corresponds to the final classification accuracy, and 2) CNN-Simple does not sufficiently exploit discriminative features. It simply combines the results from all the sampled patches, some of which might be uninformative or misleading.

LLC performs better than CNN-Simple. One of the reason is that Bag-of-Words models inherently deal with inputs with arbitrary sizes. MSPN outperforms LLC by a large margin. Apart from the reason that deep models are stronger learner, MSPN differs from LLC in that it learns task-specific filters and implicitly extracts discriminative features. Furthermore, compared to LLC, MSPN learns a much more compact image-level descriptor ($3456$ vs $12288$ dimensions), which further shows the superiority of our approach.

Figure~\ref{fig:failureCases} shows some failure cases. From it one can see that our method might fail under cases of unusual layout, blurry text and ambiguity (sometimes Japanese words are written entirely in Chinese characters and there is no way to distinguish them other than using semantics).

\subsection{Discussion}

In this section we discuss the effect of multi-stage pooling. To verify the effectiveness of multi-stage pooling, we construct several network variants by removing part of the pooling layers. Table~\ref{tbl:compVariants} shows the configurations of the variants and their corresponding recognition accuracies. We can see from the table that the performances of the pooling layers are close when they are used separately. Significant performance gain can be observed when they are combined. This demonstrate the effectiveness of multi-stage pooling.

\begin{table}[t]
\centering
\caption{Multi-stage pooling configurations for MSPN and its variants. ``\textbf{ssp-3}'' indicates that the network variant only uses spatially-sensitive pooling layer-3 (\textbf{ssp-3}) in Figure~\ref{fig:architecture}, ``\textbf{ssp-2} + \textbf{ssp-3}'' indicates the network variant that uses both \textbf{ssp-2} and \textbf{ssp-3} (*In Variant-1 the number of hidden nodes in \textbf{fc2} is set to $512$)}
\begin{tabular}{l|l|c} \hline
Variant  & Configurations & Average Error (\%)\\
\hline
Variant-1       & \textbf{ssp-1}                    & 7.3 \\
Variant-2       & \textbf{ssp-2}                    & 7.8 \\
Variant-3       & \textbf{ssp-3}*                   & 8.0 \\
Variant-4       & \textbf{ssp-2} + \textbf{ssp-3}   & 7.4 \\
Variant-5       & \textbf{ssp-1} + \textbf{ssp-2}   & 6.6 \\
\textbf{MSPN}   & \textbf{ssp-1 + ssp-2 + ssp-3}    & \textbf{5.6} \\
\hline
\end{tabular}
\label{tbl:compVariants}
\end{table}

\section{Conclusion} \label{sec:conclusion}

We have presented an effective algorithm for script identification in real-world scenarios. The proposed algorithm is able to better exploit the properties of texts in natural images. Moreover, we collected and released a large-scale benchmark for performance evaluation and comparison. The experiments on this dataset demonstrate that the proposed algorithm achieves higher performance than conventional approaches, including the original CNN method and LLC.

In this paper, we have only performed script identification in cropped word images. In future work, we plan to investigate approaches that can recognize the language type of texts from full natural images. This direction would be more promising and practical, because in reality the input to the script identification system is more likely to be full images, instead of cropped images.


\ifCLASSOPTIONcaptionsoff
  \newpage
\fi



%


\bibliographystyle{IEEEtran}
\bibliography{icdar2015siw}

\begin{thebibliography}{10}
\providecommand{\url}[1]{#1}
\csname url@samestyle\endcsname
\providecommand{\newblock}{\relax}
\providecommand{\bibinfo}[2]{#2}
\providecommand{\BIBentrySTDinterwordspacing}{\spaceskip=0pt\relax}
\providecommand{\BIBentryALTinterwordstretchfactor}{4}
\providecommand{\BIBentryALTinterwordspacing}{\spaceskip=\fontdimen2\font plus
\BIBentryALTinterwordstretchfactor\fontdimen3\font minus
  \fontdimen4\font\relax}
\providecommand{\BIBforeignlanguage}[2]{{%
\expandafter\ifx\csname l@#1\endcsname\relax
\typeout{** WARNING: IEEEtran.bst: No hyphenation pattern has been}%
\typeout{** loaded for the language `#1'. Using the pattern for}%
\typeout{** the default language instead.}%
\else
\language=\csname l@#1\endcsname
\fi
#2}}
\providecommand{\BIBdecl}{\relax}
\BIBdecl

\bibitem{DBLP:journals/pami/Tan98}
T.~N. Tan, ``Rotation invariant texture features and their use in automatic
  script identification,'' \emph{{IEEE} Trans. Pattern Anal. Mach. Intell.},
  vol.~20, no.~7, pp. 751--756, 1998.

\bibitem{DBLP:journals/pami/HochbergKTK97}
J.~Hochberg, P.~Kelly, T.~Thomas, and L.~Kerns, ``Automatic script
  identification from document images using cluster-based templates,''
  \emph{{IEEE} Trans. Pattern Anal. Mach. Intell.}, vol.~19, no.~2, pp.
  176--181, 1997.

\bibitem{DBLP:journals/pami/GhoshDS10}
D.~Ghosh, T.~Dube, and A.~P. Shivaprasad, ``Script recognition - {A} review,''
  \emph{{IEEE} Trans. Pattern Anal. Mach. Intell.}, vol.~32, no.~12, pp.
  2142--2161, 2010.

\bibitem{DBLP:conf/icdar/PhanSDLT11}
T.~Q. Phan, P.~Shivakumara, Z.~Ding, S.~Lu, and C.~L. Tan, ``Video script
  identification based on text lines,'' in \emph{In Proc. of ICDAR}, 2011.

\bibitem{DBLP:conf/das/ZhaoSLT12}
D.~Zhao, P.~Shivakumara, S.~Lu, and C.~L. Tan, ``New spatial-gradient-features
  for video script identification,'' in \emph{In Proc. of Workshop on DAS},
  2012.

\bibitem{DBLP:journals/ijdar/JoshiGS07}
G.~D. Joshi, S.~Garg, and J.~Sivaswamy, ``A generalised framework for script
  identification,'' \emph{{IJDAR}}, vol.~10, no.~2, pp. 55--68, 2007.

\bibitem{DBLP:conf/icdar/PalSC03}
U.~Pal, S.~Sinha, and B.~B. Chaudhuri, ``Multi-script line identification from
  indian document,'' in \emph{In Proc. of ICDAR}, 2003.

\bibitem{DBLP:conf/das/SinhaPC04}
S.~Sinha, U.~Pal, and B.~B. Chaudhuri, ``Word-wise script identification from
  indian documents,'' in \emph{In Proc. of Workshop on DAS}, 2004.

\bibitem{DBLP:conf/icdar/RaniDL13}
R.~Rani, R.~Dhir, and G.~S. Lehal, ``Script identification of pre-segmented
  multi-font characters and digits,'' in \emph{In Proc. of ICDAR}, 2013.

\bibitem{DBLP:conf/icdar/GomezK13}
L.~G. i~Bigorda and D.~Karatzas, ``Multi-script text extraction from natural
  scenes,'' in \emph{In Proc. of ICDAR}, 2013.

\bibitem{DBLP:conf/cvpr/WangYYLHG10}
J.~Wang, J.~Yang, K.~Yu, F.~Lv, T.~S. Huang, and Y.~Gong,
  ``Locality-constrained linear coding for image classification,'' in \emph{In
  Proc. of CVPR}, 2010.

\bibitem{DBLP:conf/nips/KrizhevskySH12}
A.~Krizhevsky, I.~Sutskever, and G.~E. Hinton, ``Imagenet classification with
  deep convolutional neural networks,'' in \emph{In Proc. of NIPS}, 2012.

\bibitem{DBLP:conf/icdar/ShahabSD11a}
A.~Shahab, F.~Shafait, and A.~Dengel, ``{ICDAR} 2011 robust reading competition
  challenge 2: Reading text in scene images,'' in \emph{In Proc. of ICDAR},
  2011.

\bibitem{DBLP:conf/eccv/WangB10}
K.~Wang and S.~Belongie, ``Word spotting in the wild,'' in \emph{In Proc. of
  ECCV}, 2010.

\bibitem{DBLP:conf/bmvc/MishraAJ12}
A.~Mishra, K.~Alahari, and C.~V. Jawahar, ``Scene text recognition using higher
  order language priors,'' in \emph{In Proc. of BMVC}, 2012.

\bibitem{lecun1998gradient}
Y.~LeCun, L.~Bottou, Y.~Bengio, and P.~Haffner, ``Gradient-based learning
  applied to document recognition,'' \emph{Proceedings of the IEEE}, vol.~86,
  no.~11, pp. 2278--2324, 1998.

\bibitem{DBLP:series/lncs/LeCunBOM12}
Y.~LeCun, L.~Bottou, G.~B. Orr, and K.~M{\"{u}}ller, ``Efficient backprop,'' in
  \emph{Neural Networks: Tricks of the Trade - Second Edition}, 2012, pp.
  9--48.

\bibitem{DBLP:conf/cvpr/YangYGH09}
J.~Yang, K.~Yu, Y.~Gong, and T.~S. Huang, ``Linear spatial pyramid matching
  using sparse coding for image classification,'' in \emph{In Proc. of CVPR},
  2009.

\bibitem{DBLP:journals/ml/CortesV95}
C.~Cortes and V.~Vapnik, ``Support-vector networks,'' \emph{Machine Learning},
  vol.~20, no.~3, pp. 273--297, 1995.

\end{thebibliography}

\end{document}